\documentclass{spie} 
\usepackage{graphicx}
\usepackage{hyperref}
\usepackage{color}
\usepackage{subcaption}
\usepackage[margin=1.5in]{geometry}

\title{Expert identification of visual primitives used by CNNs during mammogram classification}

\author[a]{Jimmy Wu}
\author[b]{Diondra Peck}
\author[c]{Scott Hsieh}
\author[d]{Vandana Dialani, MD}
\author[e]{ Constance D. Lehman, MD}
\author[a]{Bolei Zhou}
\author[f]{Vasilis Syrgkanis}
\author[f]{Lester Mackey}
\author[f]{Genevieve Patterson}
\affil[a]{CSAIL, MIT, Cambridge, USA}
\affil[b]{Harvard University, Cambridge, USA}
\affil[c]{Department of Radiological Sciences, UCLA, Los Angeles, USA}
\affil[d]{Beth Israel Deaconess Medical Center, Cambridge, USA}
\affil[e]{Massachusetts General Hospital, Cambridge, USA}
\affil[f]{Microsoft Research New England, Cambridge, USA}

\authorinfo{Corresponding authors Jimmy Wu \url{jimmywu@alum.mit.edu} and Genevieve Patterson \url{gen@microsoft.com}; This paper is being submitted solely to SPIE for publication and presentation.}

\begin{document} 
\maketitle

\begin{abstract}

This work interprets the internal representations of deep neural networks trained for classification of diseased tissue in 2D mammograms. We propose an expert-in-the-loop interpretation method to label the behavior of internal units in convolutional neural networks (CNNs). Expert radiologists identify that the visual patterns detected by the units are correlated with meaningful medical phenomena such as mass tissue and calcificated vessels. We demonstrate that several trained CNN models are able to produce explanatory descriptions to support the final classification decisions. We view this as an important first step toward interpreting the internal representations of medical classification CNNs and explaining their predictions.

\end{abstract}

\keywords{medical image understanding, deep learning for diagnosis, interpretable machine learning, expert-in-the-loop methods}

\section{Purpose}
State-of-the-art convolutional neural networks (CNNs) can now match and even supersede human performance on many visual recognition tasks~\cite{he2015delving,esteva2017dermatologist}; however, these significant advances in discriminative ability have been achieved in part by increasing the complexity of the neural network model which compounds computational obscurity~\cite{bolei2015object, szegedy2015going, He2015, xie2017aggregated}. CNN models are often criticized as black boxes because of their massive model parameters. Thus, lack of intepretability prevents CNNs from being used widely in clinical settings and for scientific exploration of medical phenomena~\cite{fenton2007influence, song2016using}.

Deep learning based cancer detection in 2D mammograms has recently achieved near human levels of sensitivity and specificity, as evidenced by the large-scale Digital Mammography DREAM Challenge.~\cite{bionetworksdigital} Recent computer aided diagnostic systems have also applied advanced machine learning to a combination of imaging data, patient demographics, and medical history with impressive results.~\cite{song2016using} However, applications such as breast cancer diagnosis and treatment heavily depend on a sense of trust between patient and practitioner, which can be impeded by black-box machine learning diagnosis systems. Thus, automated image diagnosis provides a compelling opportunity to reevaluate the relationship between clinicians and neural networks. Can we create networks that explain their decision making? Instead of producing only a coarse binary classification (e.g. does a scan show presence of disease or not), we seek to produce relevant and informative descriptions of the predictions made by a CNN in a format familiar to radiologists. In this paper, we examine the behavior of the internal representations of CNNs trained for breast cancer diagnosis. We invite several human experts to compare the visual patterns used by these CNNs to the lexicon used by practicing radiologists. We use the Digital Database for Screening Mammography (DDSM)~\cite{heath2000digital} as our training and testing benchmark. 

\subsection*{Contributions}
To our knowledge, this work is the first step toward creating neural network systems that interact seamlessly with clinicians. Our principal contributions, listed below, combine to offer insight and identify commonality between deep neural network pipelines and the workflow of practicing radiologists. Our contributions are as follows:

\begin{itemize}
\item We visualize the internal representations of CNNs trained on mammograms labeled as cancerous, normal, benign, or benign without callback
\item We develop an interface to obtain human expert labels for the visual patterns used by CNNs in cancer prediction
\item We compare the expert-labeled internal representations to the BI-RADS lexicon~\cite{reporting1998data}, showing that many interpretable internal CNN units detect meaningful factors used by radiologists for breast cancer diagnosis
\end{itemize}

\section{Methods}
To gain a richer understanding of which visual primitives CNNs use to predict cancer, we fine-tuned several strongly performing networks on training images from the Digital Database for Screening Mammography (DDSM).~\cite{heath2000digital} For each fine-tuned network, we then evaluated the visual primitives detected by the individual units using Network Dissection, a technique to visualize the favorite patterns detected by each unit.~\cite{bau2017network} Three authors who are practicing radiologists or experts in this area manually reviewed the unit visualizations and labeled the phenomena identified by each unit. Finally, we compared the named phenomena used by the CNN internal units to items in the BI-RADS lexicon~\cite{reporting1998data}. Note that we denote the convolutional filters at each layer as units, as opposed to 'neurons', to disambiguate them from the biological entities.

\subsection{Dataset} 
We conduct our experiments with images from the Digital Database for Screening Mammography (DDSM), a dataset compiled to facilitate research in computer-aided breast cancer screening. DDSM consists of 2,500 studies, each including two images of each breast, patient age, ACR breast density rating, subtlety rating for abnormalities, ACR keyword description of abnormalities, and information about the imaging modality and resolution. Labels include image-wide designations (e.g., cancerous, normal, benign, and benign without callback) and pixel-wise segmentations of lesions.~\cite{heath2000digital}

For the experiments in the following sections, we divided the DDSM dataset scans into 80\% train, 10\% validation, and 10\% test partitions. All images belonging to a unique patient are in the same split, to prevent training and testing on different views of the same breast. 

\subsection{Network Architectures} 
We adapted several well-known image classification networks for breast cancer diagnosis as shown in Table~\ref{tab:architectures}. We modified the final fully connected layer of each architecture to have two classes corresponding to a positive or negative diagnosis. Network weights were initialized using the corresponding pretrained ImageNet~\cite{deng2009imagenet} networks and fine-tuned on DDSM. We trained all networks in the PyTorch~\cite{paszke2017pytorch} framework using stochastic gradient descent (SGD) with learning rate 0.0001, momentum 0.9, and weight decay 0.0001.

\begin{table}[h]
\begin{center}
\begin{tabular}{ |c|c| } 
 \hline
 Architecture & AUC \\
 \hline
 AlexNet~\cite{krizhevsky2012imagenet}  & 0.8632 \\ 
 VGG-16~\cite{simonyan2014very}  & 0.8929 \\ 
 Inception-v3~\cite{szegedy2015going} & 0.8805 \\ 
 ResNet-152~\cite{He2015}  & 0.8757 \\ 
 \hline
\end{tabular}
\vspace{3mm}
\caption{The network architectures used and their performance as the AUC on the validation set.}
\label{tab:architectures}
\end{center}
\end{table}

\begin{figure}[h!]
\centering
\includegraphics[scale=.3]{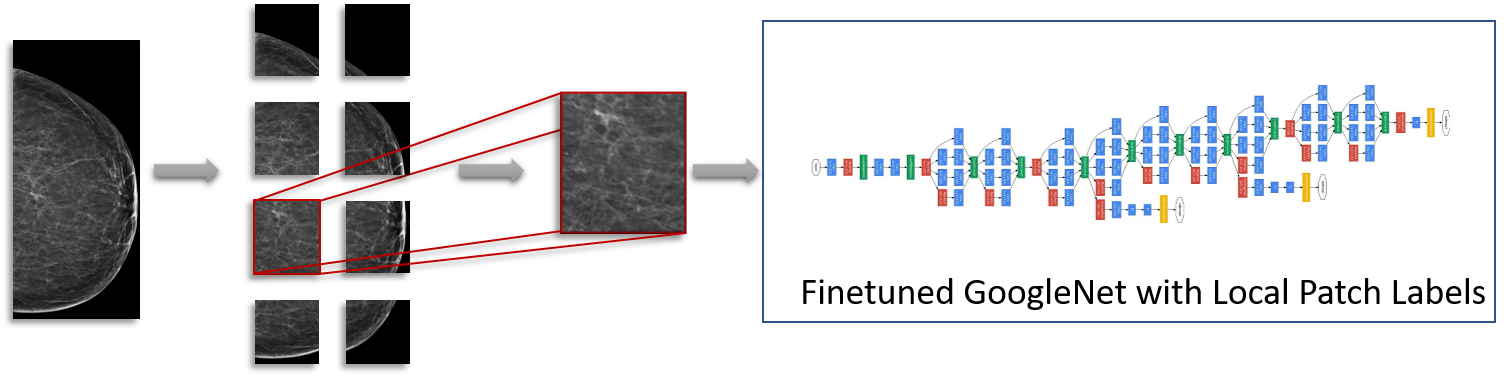}
\vspace{3mm}
\caption{GoogleNet Inception-v3 architecture fine-tuned with local image patches and their labels. Multiple overlapping patches are extracted from each image with a sliding window and then passed through a CNN with the local patch label determined by the lesion masks from DDSM. After fine-tuning each network we tested performance on the task of classifying whether the patch contains a malignant lesion. We show the performance of our various fine-tuned networks in Table ~\ref{tab:architectures}.} 
\label{fig:patch-arch}
\end{figure}

Figure~\ref{fig:patch-arch} illustrates how we prepared each mammogram for training and detection. Because of the memory requirements of processing a high-resolution image with a CNN, we split the mammograms into patches and process image patches. We applied a sliding window at 25\% the size of a given mammogram with a 50\% patch stride. This gave us a set of image patches for each mammogram, a subset of which may not have contained any cancerous lesions. The ground truth label for each mammogram patch was computed as positive for cancer if at least 30\% of a cancerous lesion was contained in the image patch or at least 30\% of the image patch was covered by a lesion; all other patches were assigned a negative label. Lesion locations were determined from the lesion segmentation masks of DDSM.

\subsection{Network Dissection} 
\label{sec:netdis}
Network Dissection (NetDissect) is a recent method proposed for assessing how well visual concepts are disentangled within CNNs.~\cite{bau2017network} Network Dissection defines and quantifies the intepretability as a measure of how well individual units align with sets of human-interpretable concepts. 

\begin{figure}[h!]
    \centering
    \begin{subfigure}[b]{\textwidth}
        \includegraphics[width=\textwidth]{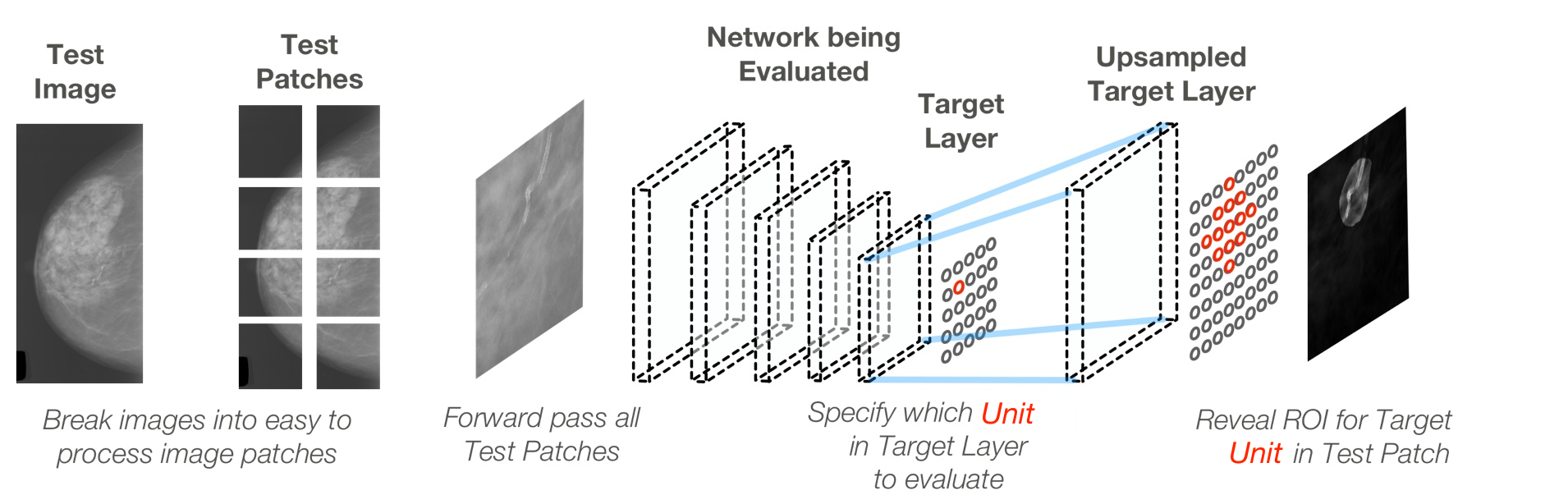}        
	\caption{Illustration of how Network Dissection proceeds for a single instance. Above, one unit is probed to display the Region of Interest (ROI) in the evaluated image responsible for that unit's activation value. ROIs may not line up directly as shown in this figure, please see Bau et al.~\cite{bau2017network} for a complete description of this process.}
        \label{fig:netdissect_one}
    \end{subfigure}
    \begin{subfigure}[b]{\textwidth} 
        \includegraphics[width=\textwidth]{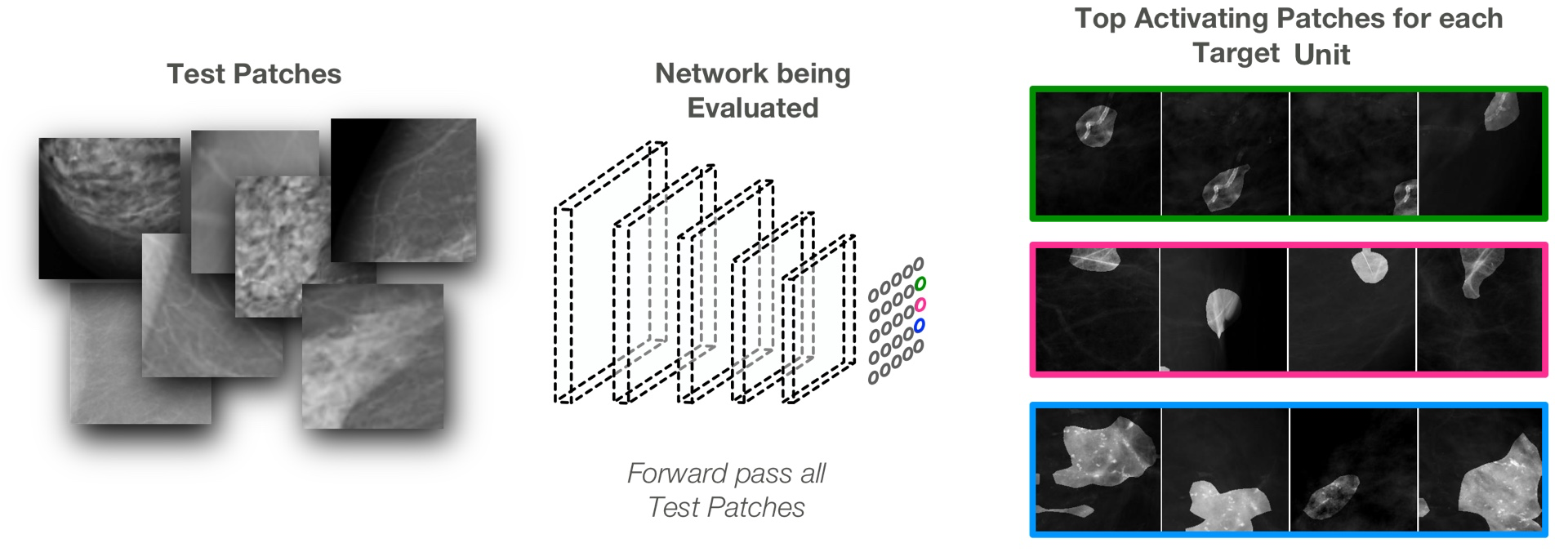}
	\caption{Illustration of how Network Dissection proceeds for all units of interest in a given convolutional layer. All images from a test set are processed in the manner of Fig.~\ref{fig:netdissect_one}. The top activating test images for each unit are recorded to create a visualization of the unit's top activated visual phenomena. Each top activating image is segmented by the upsampled and binarized feature map of that unit.}
	\label{fig:netdissect_multi}
    \end{subfigure}       
    \vspace{1.5mm}
    \caption{Illustration of Network Dissection for identifying the visual phenomena used by a CNN of interest.}
\label{fig:netdis}
\end{figure}

Figure~\ref{fig:netdis} demonstrates at a high level how NetDissect works to interpret the units at a target layer of a network. We used our validation split of DDSM to create visualizations for the final convolutional layer units of each evaluated network. For ResNet-152, we also evaluated the second to last convolutional layer due to its high network depth. Because of the hierarchical structure of CNNs, the final convolutional layer will contain the most high-level semantic concepts, whereas the earlier layers will contain mostly low-level gradient features. We choose to evaluate the final convolutional layers containing high-level semantic features, as they are more likely to be aligned with the visual taxonomy used by radiologists. Note that the NetDissect approach to unit visualization applies only to convolutional network layers due to their maintenance of spatial information.

Figure~\ref{fig:netdissect_multi} shows how we created the unit visualizations for our analysis in Sections~\ref{sec:expert_eval} and~\ref{sec:results}. We passed all image patches in our validation set through each of our four networks. For each unit in the target convolution layer being evaluated, we recorded the unit's maximum activation value, denoted as the unit's score, as well as the ROI from the image patch that caused the measured activation. To visualize each unit (Figs.~\ref{fig:survey} and~\ref{fig:vis_units}), we display the top activating image patches in order sorted by their score for that unit. Each top activating image is further segmented by the upsampled and binarized feature map of that unit to highlight the highly activated image region.

\subsection{Human Evaluation of Visual Primitives used by CNNs}
\label{sec:expert_eval}

\begin{figure}[h!]
    \centering 
    \includegraphics[width=\textwidth]{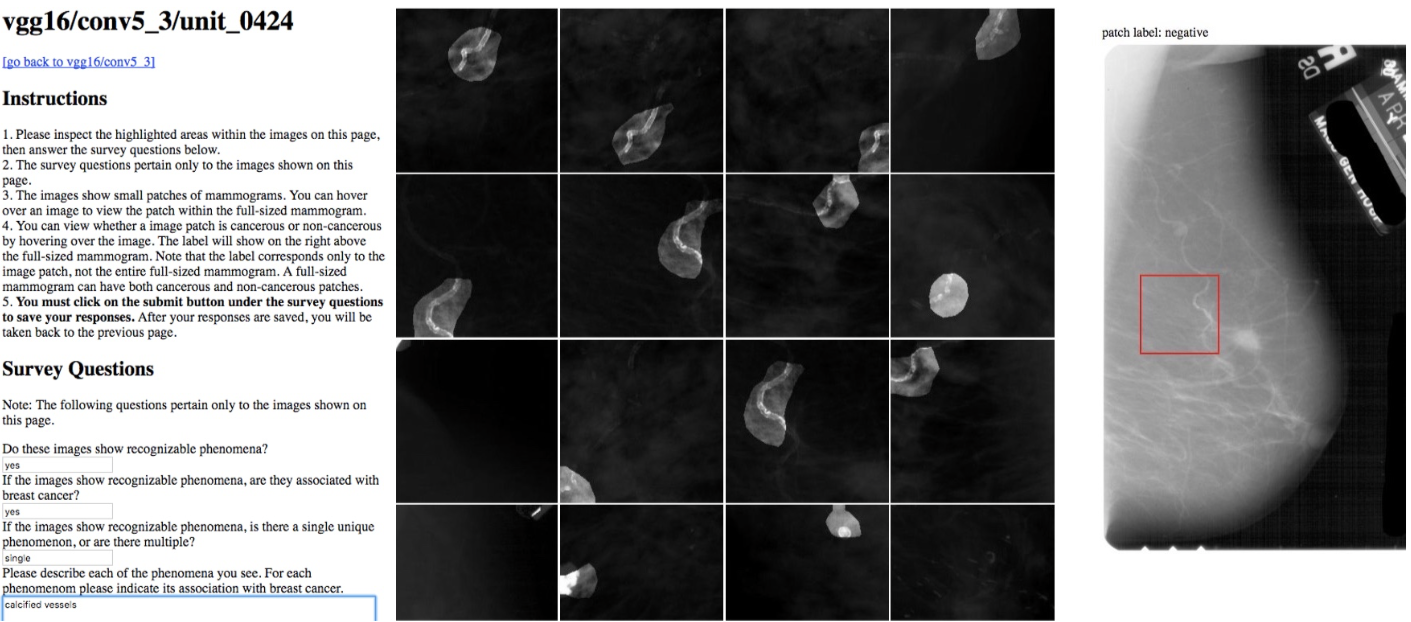}
    \vspace{3mm}
    \caption{ Web-based Survey Tool: This user interface was used to ask the expert readers about units of interest. The survey asked questions such as: ``Do these images show recognizable phenomena?'' and ``Please describe each of the phenomena you see. For each phenomenon please indicate its association with breast cancer.'' In the screenshot above, one expert has labeled the unit's phenomena as `Calcified Vessels'.}
    \label{fig:survey}
\end{figure}

To further validate the visual primitives discovered by our networks, we created a web-based survey to solicit input from expert readers. The expert readers consisted of two radiologists specialized in breast imaging and one medical physicist. A screenshot from the survey tool is shown in Figure~\ref{fig:survey}.  The survey provided a list of 40 to 50 units culled from the final layer of one of our neural networks. The neural network often had many more units, too many for exhaustive analysis with three expert readers. Thus, the units that were selected were composed partly of the top activating patches that all or mostly contained cancer and partly of a random selection of other patches.

The readers were able to see a preview of each unit, which consisted of several image patches with highlights indicating the regions of interest that caused the unit to activate most strongly. From this preview, the readers were able to formulate an initial hypothesis of what each unit is associated with. The readers could click on any preview to select a specific unit to focus on, and would then be brought to a second page dedicated specifically to that unit, showing additional patches as well as the context of the entire mammogram, as shown in Figure~\ref{fig:survey}. On this page, users could then comment on the unit in a structured report, indicating if there was any distinct phenomenon associated with the unit, and if so, its relationship to breast cancer. The web-based survey saved results after each unit and could be accessed over multiple sessions to avoid reader fatigue.

Some of the units shown had no clear connection with breast cancer and would appear to be spurious. Still other units presented what appeared to be entangled events, such as mixtures of mass and calcification, that were associated with malignancy but in a clearly identifiable way. However, many of the units appeared to show a clean representation of a single phenomenon known to be associated with breast cancer.

\section{Results}
\label{sec:results}
We compared the expert-annotated contents of 134 units from four networks to the lexicon of the BI-RADS taxonomy.~\cite{reporting1998data, fenton2007influence} This qualitative evaluation was designed to estimate the overlap between the standard system used by radiologists to diagnose breast cancer and the visual primitives used by our trained CNNs. 

\begin{figure}[th!]%
    \centering 
    \includegraphics[width=\textwidth]{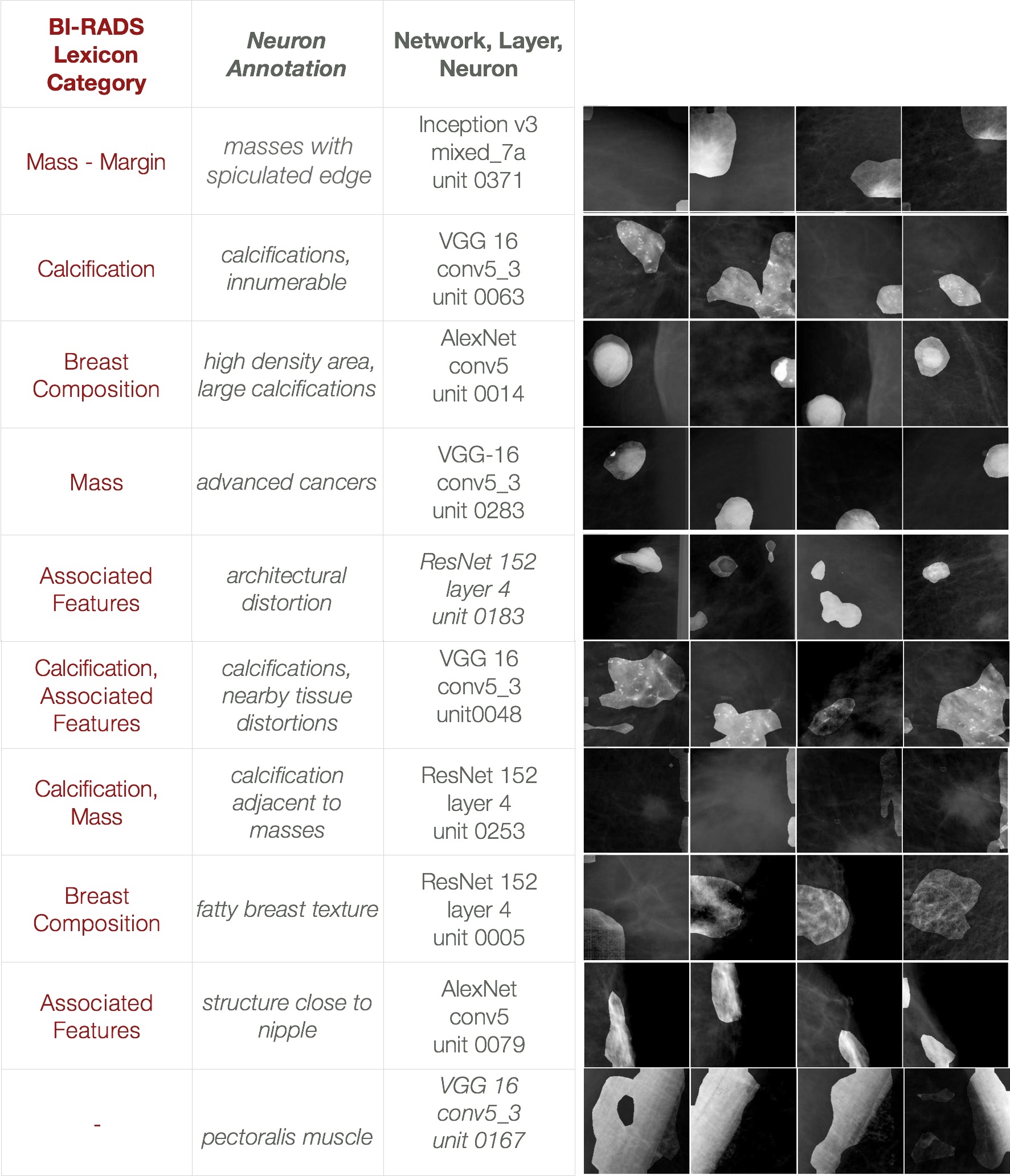}
    \vspace{3mm}
    \caption{The table above shows some of the labeled units and their interpretations. The first column lists the general BI-RADS category associated with the units visualized in the last column. The second column displays the expert annotation of the visual event identified by each unit, summarized for length. The third column lists the network, convolutional layer, and the unit's ID number.}

    \label{fig:vis_units}
\end{figure}

Direct classification of BI-RADS entities has long been a topic of interest in machine learning for mammography.~\cite{orel1999bi} Our experiments differ from direct classification because our training set was constructed with simple positive/negative labels instead of detailed BI-RADS categories. In this work we chose a well-understood medical event, the presence of cancer in mammograms, to evaluate if unit visualization is a promising avenue for discovering important visual phenomena in less well-understood applications. Our results, shown in Fig.~\ref{fig:vis_units}, show that networks trained to recognize cancer end up using many of the BI-RADS categories even though the training labels given to the network simply indicated the presence or absence of cancer.

Units in all networks identify advanced cancers, large benign masses, and several kinds of obvious calcifications. Encouragingly, many units also identify important associated features such as spiculation, breast density, architectural distortions, and the state of tissue near the nipple. Several units in Fig.~\ref{fig:vis_units} show that the CNNs use breast density and parenchymal patterns to make predictions. This network behavior could be used to find a new computational perspective on the relationship between breast density, tissue characteristics, and cancer risk, which has been a popular research topic for the last 25 years.~\cite{oza1993mammographic, petroudi2003automatic, mccormack2006breast} 

\section{Conclusion}
In this exploratory study, we trained CNNs for classification of diseased tissue in mammograms and investigated the visual concepts used by the internal units of the CNNs. Using an expert-in-the-loop method, we discovered that many CNN units identify recognizable medical phenomena used by radiologists. Indeed, Fig.~\ref{fig:vis_units} shows significant overlap with the BI-RADS lexicon. We note however, that some units had no identified connection with breast cancer, and yet other units identified entangled events. We believe these findings are an important first step towards interpreting the decisions made by CNNs in classification of diseased tissue.


\bibliography{references}

\begin{thebibliography}{10}

\bibitem{he2015delving}
He, K., Zhang, X., Ren, S., and Sun, J., ``Delving deep into rectifiers:
  Surpassing human-level performance on imagenet classification,'' in [{\em
  ICCV}{\nolinebreak\hspace{0.1em}]},   1026--1034 (2015).

\bibitem{esteva2017dermatologist}
Esteva, A., Kuprel, B., Novoa, R.~A., Ko, J., Swetter, S.~M., Blau, H.~M., and
  Thrun, S., ``Dermatologist-level classification of skin cancer with deep
  neural networks,'' {\em Nature}~{\bf 542}(7639),  115--118 (2017).

\bibitem{bolei2015object}
Bolei, Z., Khosla, A., Lapedriza, A., Oliva, A., and Torralba, A., ``Object
  detectors emerge in deep scene cnns,'' {\em ICLR}  (2015).

\bibitem{szegedy2015going}
Szegedy, C., Liu, W., Jia, Y., Sermanet, P., Reed, S., Anguelov, D., Erhan, D.,
  Vanhoucke, V., and Rabinovich, A., ``Going deeper with convolutions,'' in
  [{\em Computer Vision and Pattern Recognition
  (CVPR)}{\nolinebreak\hspace{0.1em}]},  (2015).

\bibitem{He2015}
He, K., Zhang, X., Ren, S., and Sun, J., ``Deep residual learning for image
  recognition,'' {\em arXiv preprint arXiv:1512.03385}  (2015).

\bibitem{xie2017aggregated}
Xie, S., Girshick, R., Doll{\'a}r, P., Tu, Z., and He, K., ``Aggregated
  residual transformations for deep neural networks,'' in [{\em Computer Vision
  and Pattern Recognition (CVPR), 2017 IEEE Conference
  on}{\nolinebreak\hspace{0.1em}]},   5987--5995, IEEE (2017).

\bibitem{fenton2007influence}
Fenton, J.~J., Taplin, S.~H., Carney, P.~A., Abraham, L., Sickles, E.~A.,
  D'orsi, C., Berns, E.~A., Cutter, G., Hendrick, R.~E., Barlow, W.~E., et~al.,
  ``Influence of computer-aided detection on performance of screening
  mammography,'' {\em New England Journal of Medicine}~{\bf 356}(14),
  1399--1409 (2007).

\bibitem{song2016using}
Song, L., Hsu, W., Xu, J., and Van Der~Schaar, M., ``Using contextual learning
  to improve diagnostic accuracy: Application in breast cancer screening,''
  {\em IEEE journal of biomedical and health informatics}~{\bf 20}(3),
  902--914 (2016).

\bibitem{bionetworksdigital}
Bionetworks, S., ``Digital mammography dream challenge,'' (2016).

\bibitem{heath2000digital}
Heath, M., Bowyer, K., Kopans, D., Moore, R., and Kegelmeyer, W.~P., ``The
  digital database for screening mammography,'' in [{\em Proceedings of the 5th
  international workshop on digital mammography}{\nolinebreak\hspace{0.1em}]},
   212--218, Medical Physics Publishing (2000).

\bibitem{reporting1998data}
Reporting, B.~I., ``Data system (bi-rads),'' {\em Reston VA: American College
  of Radiology}  (1998).

\bibitem{bau2017network}
Bau, D., Zhou, B., Khosla, A., Oliva, A., and Torralba, A., ``Network
  dissection: Quantifying interpretability of deep visual representations,''
  {\em CVPR}  (2017).

\bibitem{deng2009imagenet}
Deng, J., Dong, W., Socher, R., Li, L.-J., Li, K., and Fei-Fei, L., ``Imagenet:
  A large-scale hierarchical image database,'' in [{\em Computer Vision and
  Pattern Recognition, 2009. CVPR 2009. IEEE Conference
  on}{\nolinebreak\hspace{0.1em}]},   248--255, IEEE (2009).

\bibitem{paszke2017pytorch}
Paszke, A., Gross, S., Chintala, S., and Chanan, G., ``Pytorch,'' (2017).

\bibitem{krizhevsky2012imagenet}
Krizhevsky, A., Sutskever, I., and Hinton, G.~E., ``Imagenet classification
  with deep convolutional neural networks,'' in [{\em Advances in neural
  information processing systems}{\nolinebreak\hspace{0.1em}]},   1097--1105
  (2012).

\bibitem{simonyan2014very}
Simonyan, K. and Zisserman, A., ``Very deep convolutional networks for
  large-scale image recognition,'' {\em arXiv preprint arXiv:1409.1556}
  (2014).

\bibitem{orel1999bi}
Orel, S.~G., Kay, N., Reynolds, C., and Sullivan, D.~C., ``Bi-rads
  categorization as a predictor of malignancy,'' {\em Radiology}~{\bf 211}(3),
  845--850 (1999).

\bibitem{oza1993mammographic}
Oza, A.~M. and Boyd, N.~F., ``Mammographic parenchymal patterns: a marker of
  breast cancer risk.,'' {\em Epidemiologic reviews}~{\bf 15}(1),  196--208
  (1993).

\bibitem{petroudi2003automatic}
Petroudi, S., Kadir, T., and Brady, M., ``Automatic classification of
  mammographic parenchymal patterns: A statistical approach,'' in [{\em
  Engineering in Medicine and Biology Society, 2003. Proceedings of the 25th
  Annual International Conference of the IEEE}{\nolinebreak\hspace{0.1em}]},
  {\bf 1},  798--801, IEEE (2003).

\bibitem{mccormack2006breast}
McCormack, V.~A. and dos Santos~Silva, I., ``Breast density and parenchymal
  patterns as markers of breast cancer risk: a meta-analysis,'' {\em Cancer
  Epidemiology and Prevention Biomarkers}~{\bf 15}(6),  1159--1169 (2006).

\end{thebibliography}
\bibliographystyle{spiebib} 

\end{document}